\pdfoutput=1
\documentclass[11pt]{article}

\usepackage{acl}
\usepackage{times}
\usepackage{latexsym}
\usepackage[T1]{fontenc}
\usepackage[utf8]{inputenc}

\usepackage{microtype}

\usepackage{url}
\usepackage{booktabs}
\usepackage{multirow}
\usepackage{array, caption, floatrow, makecell, booktabs}
\usepackage{wrapfig}
\usepackage{floatrow}
\usepackage{comment}
\usepackage{xspace,mfirstuc,tabulary}
\usepackage{graphicx}
\usepackage{stfloats,framed}
\usepackage{lipsum}
\usepackage{amsmath}
\usepackage{amsfonts}
\usepackage{amssymb}
\usepackage{enumitem}
\usepackage{xcolor}

\usepackage{subcaption}
\definecolor{ballblue}{rgb}{0.0, 0.53, 0.74}
\definecolor{crimson}{rgb}{0.86, 0.08, 0.24}
\definecolor{dark}{rgb}{0.4, 0.4, 0.4}
\newcommand{\our}{SCoTD\xspace}

\newcommand\blfootnote[1]{%
  \begingroup
  \renewcommand\thefootnote{}\footnote{#1}%
  \addtocounter{footnote}{-1}%
  \endgroup
}
\title{Symbolic Chain-of-Thought Distillation: \\ Small Models Can Also ``Think'' Step-by-Step}
\author{Liunian Harold Li$^{*\dagger}$, %\thanks{\,\,\,Work done during an internship at AI2.}, 
Jack Hessel$^{\clubsuit}$, Youngjae Yu$^{\diamondsuit}$, \\ \textbf{ Xiang Ren$^\circ$, Kai-Wei Chang$^\dagger$ \& Yejin Choi$^\clubsuit$$^\heartsuit$}\\
$^\dagger$University of California, Los Angeles,
$^\clubsuit$Allen Institute for Artificial Intelligence\\ 
$^\circ$University of Southern California, 
$^\diamondsuit$ Yonsei University,
$^\heartsuit$University of Washington \\
}

\begin{document}
\maketitle

\begin{abstract}

Chain-of-thought prompting (e.g., ``Let's think step-by-step") primes large language models to verbalize rationalization for their predictions. While chain-of-thought can lead to dramatic performance gains, benefits appear to emerge only for sufficiently large models (beyond 50B parameters). We show that orders-of-magnitude smaller models (125M---1.3B parameters) can still benefit from chain-of-thought prompting. To achieve this, we introduce \emph{Symbolic Chain-of-Thought Distillation} (SCoTD), a method to train a smaller student model on rationalizations sampled from a significantly larger teacher model. Experiments across several commonsense benchmarks show that: 1) SCoTD enhances the performance of the student model in both supervised and few-shot settings, and especially for challenge sets; 2) sampling many reasoning chains per instance from the teacher is paramount; and 3) after distillation, student chain-of-thoughts are judged by humans as comparable to the teacher, despite orders of magnitude fewer parameters. We test several hypotheses regarding what properties of chain-of-thought samples are important, e.g., diversity vs. teacher likelihood vs. open-endedness. We release our corpus of chain-of-thought samples and code.

\end{abstract}

\section{Introduction}

\begin{figure}[t]
\centering
\includegraphics[width=\textwidth]{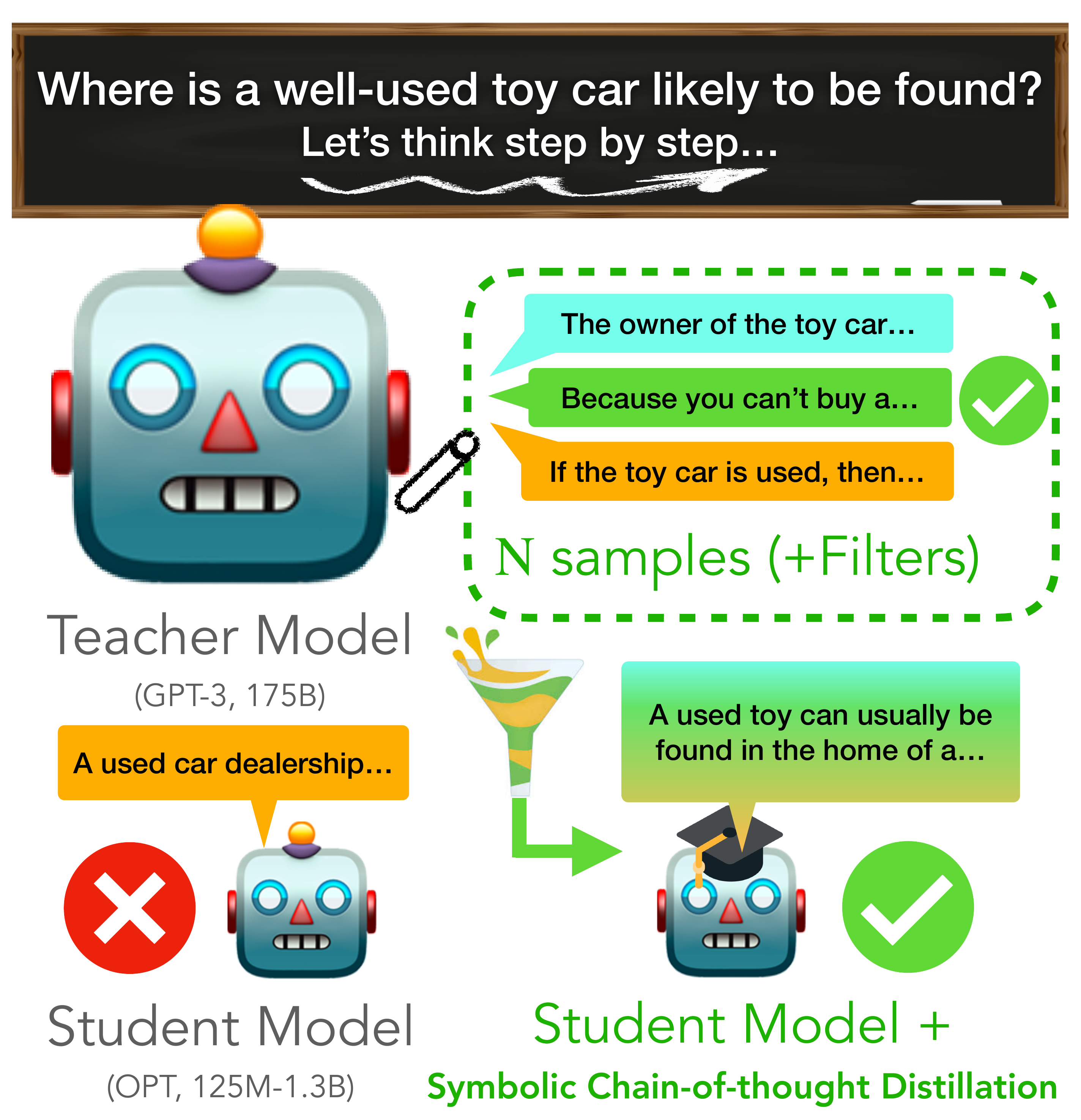}
\caption{Symbolic Chain-of-thought Distillation (SCoTD) applied to a student model, ranging in size from 125M--1.3B parameters. We show that fine-tuning on a (diverse and potentially filtered) corpus of expert chain-of-thought demonstrations from a teacher model is an effective strategy to make smaller models capable of chain-of-thought reasoning.}
\label{fig:lead_figure}
\end{figure}

\blfootnote{*Work done during an internship at AI2.}
Empirical scaling laws suggest that the accuracy of Large Language Models (LLMs) on benchmark tasks can be improved by increasing model size and pre-training data volume \cite{hoffmann2022training}. Beyond these training-time improvements, however, an inference-time strategy dubbed ``chain-of-thought" (CoT) prompting,\footnote{Sometimes called ``self-rationalization" or ``prompting with explanations.'' We will use these terms interchangeably in this paper.} i.e., eliciting verbalizations of predictive processes via key-phrases like ``Let's think step-by-step" \cite{kojima2022large}, can similarly improve performance, e.g.,  \citet{suzgun2022challenging} demonstrate additional performance gains on a hard subset of the BigBench tasks \cite{bigbench} using chain-of-thought.

% have demonstrated strong performance on few-shot learning

%can benefit from ``prompting with explanations'' and ``self-rationalize''. Recent work shows that large language models can be prompted to provide explanations.

However, chain-of-thought prompting has only been shown to be beneficial for models of sufficient scale (e.g., with more than 60B parameters ~\cite{wei2022chain}). % For smaller models, performance degradation when using this strategy is often observed. \harold{I am not sure if our results support this. In fact, in our results (Table \ref{table:main}), the small OPT is so bad at in-context learning, it never achives above random change; there is not much of ``performance degradation (or increase)''.} % One hypothesis is that capacity for self-rationalization is %emergent in large language models as a result of ``an implicit process of multitask learning'' \cite{T0}. It is of great interest whether neural models can learn to rationalize their predictions and whether learning with explanations help the model make better predictions. 
In this work, we study whether small language models can be ``taught" the capacity for chain-of-thought reasoning %via distillation from 
by larger language models. % \harold{I am not sure if we want to differentiate between 1) the ability to generate chain-of-thought and 2) the ability to perform chain-of-thought reasoning for novel tasks with a few-shot CoT prompt. Not sure if we say: `` chain-of-thought prompting has only been shown to be beneficial'', people might assume that we focuse on enalble 2 for smaller models (though we do have a preliminary experiment in Sec 3.3).} % --- JMH: I updated wording in next paragraph to highlight these results
% We are inspired by several lines of research: 1) learning with explanations; 2) recent studies that show large language models can be prompted to generate good rationales; 3) a growing number of studies that study the possibility of using machine-generated corpus as training data.
We adopt a simple strategy, which we call \emph{Symbolic Chain-of-thought Distillation} (SCoTD): first, we sample chain-of-thought rationales from large language model given (unlabeled) input instances from a dataset; then, we train a smaller language model to predict the sampled rationale and sampled label. This process follows the ``symbolic knowledge distillation'' paradigm as in \citet{west2021symbolic}, wherein corpora are sampled from a larger language model to serve as training data for a smaller one.%\cite{xiong2019pretrained,petroni2019language,schick-schutze-2021-generating,west2021symbolic,liu2022wanli,bhagavatula2022i2d2}. %See Figure~\ref{fig:lead_figure} and Table~\ref{tab:example_CoTs} for an illustration and examples.
% XIANG: beef up on 

\begin{table*}[ht]
    \centering
    \resizebox{0.97\textwidth}{!}{%
    \footnotesize
    \begin{tabular}{p{.35\linewidth}p{.65\linewidth}}
        
        \toprule
        \underline{CommonsenseQA}:
        Q: The man working in the attic swatted away a bee, but soon the single bee was an entire what? \newline
        (a) \textcolor{ballblue}{swarm}; 
        (b) pack;
        (c) countryside;
        (d) soft drink;
        (e) field of flowers
         &
        \textcolor{dark}{\emph{GPT-3:}}
        The answer must be related to bees, but also connected with being numerous. Of the above answers, only swarm fits the scenario. So the answer is: \textcolor{ballblue}{(a)} \vspace{1mm} \newline
         \textcolor{dark}{\emph{OPT-1.3B:}}
        The answer must be a swarm of bees. Of the above choices, only soft drink is used to describe a swarm of bees. So the answer is: \textcolor{crimson}{(c)}  \vspace{1mm} \newline
        \textcolor{dark}{ \emph{OPT-1.3B + SCoTD:}} The answer must be the name of something that has bees. Of the above choices, only swarms have bees. So the answer is: \textcolor{ballblue}{(a)} \\
        \midrule

        \underline{QuaRel}:
        Q: Steve is waiting at the gym ... %for his brother's basketball practice to end.
        He starts hitting around a hockey puck with his stick. The gym has a carpeted floor and more friction than the ice rink he goes to. 
        
        Which surface is rougher? (A) \textcolor{ballblue}{carpet} (B) ice rink
         &
         \textcolor{dark}{\emph{GPT-3:}}
        When something is smoother, it is easier to slide on and easier to pass through. So the carpet is rougher. So the answer is: \textcolor{ballblue}{(A)} \vspace{1mm} 
        \newline
         \textcolor{dark}{\emph{OPT-1.3B:}}
        \_\_\_\_\_\_\_\_\_ is rougher than carpet. Thus, the gym floor is rougher than the ice rink. So the answer is: \textcolor{ballblue}{(A)} \vspace{1mm} 
        \newline
         \textcolor{dark}{\emph{OPT-1.3B + SCoTD:} }
        When something is rougher, it has more friction. Thus, the gym has more friction than the ice rink he goes to. So the answer is: \textcolor{ballblue}{(A)}
        \\
        \midrule

        \underline{OpenBookQA}:
        Q: Magnets may be attracted to 
        
        (a) clay pots (b) \textcolor{ballblue}{roofing nails} (c) paper plates (d) plastic cutlery
         &
         \textcolor{dark}{\emph{GPT-3:}}
        Magnets are attracted to metal objects. These objects include roofing nails. So the answer is: \textcolor{ballblue}{(b)} \vspace{1mm} 
        \newline
         \textcolor{dark}{\emph{OPT-1.3B:}}
        Magnets are attracted to clay pots, roofing nails, paper plates, plastic cutlery. So the answer is: \textcolor{crimson}{(d)} \vspace{1mm} 
        \newline
         \textcolor{dark}{\emph{OPT-1.3B + SCoTD:} }
        Magnets may be attracted to some metals, but not to clay pots, roofing nails, paper plates or plastic cutlery. So the answer is: \textcolor{ballblue}{(b)}
        \\

        \bottomrule
    \end{tabular}%
    }
    \caption{Few-shot chain-of-thoughts produced by \emph{GPT-3} (\texttt{code-davinci-002}, the teacher model), \emph{OPT-1.3B} (the un-distilled student model), and \emph{OPT-1.3B + SCoTD} (ours), the student model trained using Symbolic Chain-of-thought Distillation. \S\ref{sec:sec_with_experiments} shows this process significantly improves the task-accuracy of the student model in a variety of settings, and in \S\ref{sec:sec_with_human_experiments}, human evaluations show that, even when the un-distilled student model happens to get the multiple choice question correct (see QuaRel example), humans tend to prefer \emph{OPT-1.3B + SCoTD}.}
    \label{tab:example_CoTs}
\end{table*}

We find that through SCoTD, smaller language models learn to self-rationalize and perform significantly better on 3 commonsense QA tasks compared to learning without rationalizations. This result holds for both supervised and few-shot settings, and across student models of varying scales (125M--1.3B parameters).
Performance gains are especially pronounced when applying distilled chain-of-thought models to difficult scenarios like: contrast sets \cite{gardner2020evaluating} (\S\ref{sec:sec_with_contrast_sets}; SCoTD significantly outperforms supervised learning on labels) and fully held-out tasks (\S\ref{sec:sec_with_fully_held_out_task}; few-shot SCoTD significantly outperforms in-context learning). % \harold{Do we want to mention the results of Sec 3.3? E.g., The student model can also generate chain-of-thoughts for an unseen task with a few demonstrations.}  Yes. -- jmh

Key to the success of this process is sampling a relatively large number of rationales per example from the teacher model (e.g., 30 rationales/example) (Figure \ref{fig:main_result}). This is different from many prior practices that train with one rationale per example~\cite{camburu2018snli,li2022explanations}. %\yj{I may be wrong, but self-consistency sample different multiple rationale?!,so this could mislead when they're not familiar with this work, provide -> train? any idea  } % In fact, this observation is closely related to the ``self-consistency'' observation ~\citep{wang2022self}, where 1) the large language models ration the probability between different valid and diverse rationales, 2) the probability assigned by the large language model may only loosely correlate with the label correctness, and 3) a marginalization over all possible rationales (approximated with sampling and then majority vote) is required to uncover the true label predicted by the language model. \harold{Will go back to work on this later}
% \harold{Not sure if we could move some discussion from ``Self-consistency in student models.'' (Sec 3) here or in the Sec 2.} \jack{lets keep this more subtle point for later.}
In ablation studies, we investigate several competing hypotheses for what are the most important factors within the corpus: we filter the corpus to CoTs that are assigned \textit{high probability} by GPT-3 %\footnote{Our notion of correctness is related to `self-consistency'' ~\citep{wang2022self} where the large language models ration the probability between different valid and diverse rationales. \jack{actually, I am not sure how this relates, can you take a look harold? I think there's a rhetorical point we can make with this citation in this section}}
vs. filtering to CoTs that are \textit{diverse} vs. filtering to CoTs that explain more \textit{open-ended} input instances. While diversity and high probability are reasonable filters that on average perform well, the ``null hypothesis'' of random downsampling performs well, suggesting that the sheer volume of the rationales is also a key contributing factor.

We release code and the corpus of sampled chain-of-thoughts at \url{https://github.com/liunian-harold-li/scotd}.

\section{Symbolic Chain-of-Thought Distillation}
\label{sec:approach}
Our primary goal is to improve the accuracy of a (relatively small) student language model $\mathcal{S}$ on a target classification\footnote{Future work would be well suited to consider if chain-of-thought prompting can be useful for generative tasks.} task $\mathcal{D}_{\texttt{Test}} = \{(x_i, y_i)\}$.\footnote{In practice, we primarily consider CommonsenseQA~\cite{talmor2019commonsenseqa}, OpenBookQA~\cite{OpenBookQA2018}, and QuaRel~\cite{tafjord2019quarel} as $\mathcal{D}$.} We assume access to 1) (an unlabeled) training set $\mathcal{D}_{\texttt{Train}} = \{(x_i)\}$; and 2) a large teacher language model $\mathcal{T}$ (e.g., GPT-3~\cite{NEURIPS2020_1457c0d6}), capable of generating chain-of-thoughts in a few-shot fashion.

Our first step is to curate a set of labeled chain-of-thoughts to serve as few-shot $\mathcal{P}$rompts for $\mathcal{T}$. For each target task, we sample a small number (e.g., 10) of examples $x_i$ from $\mathcal{D}_{\texttt{Train}}$, provide a gold classification label $y_i$, and manually author a chain-of-thought $z_i$ for each to form the prompt set $ \mathcal{P} = \{(x_i, y_i, z_i)\} $\footnote{In addition to authoring our own, we reuse chain-of-thought prompts from prior work~\citep{wei2022chain,wang2022self} when available.}.

% few-shot training set $\mathcal{D}_{\texttt{Train}}$, which contains a small number of examples, e.g., 10. %label examples from this set $(x_i, y_i)$ and
Then, for each $x_i$ in $\mathcal{D}_{\texttt{Train}}$, we sample $N$ chain-of-thoughts $\tilde z_i$ along with the resulting prediction $\tilde y_i$ from the teacher model, i.e.,
$$
(\tilde y^k_i, \tilde z^k_i) \sim_N \mathcal{T}(y_i, z_i | x_i, \mathcal{P}).
$$
% where $p_j = (x_j, y_j, z_j)$ are from $\mathcal{P}$. %\jack{how many do we put in the prompt when sampling?} \harold{We actually used all P.}
The result of this sampling is a corpus $\mathcal{C} = \{ (x_i, \{(\tilde y^k_i, \tilde z^k_i)\}^N_{k=1} ) \}$, which contain teacher-predicted chain-of-thoughts/labels. Depending on the experimental setting (details in \S~\ref{sec:sec_with_experiments}), we sometimes filter the entries of $\mathcal{C}$, e.g., in the fully supervised case where $\mathcal{D}_{\texttt{Train}}$ instances have associated labels, we discard samples for which the sample the teacher model predicted an incorrect label.
Next, we train the student model using the standard language modeling loss, i.e., we maximize
$$
E_{(x, \tilde{y}, \tilde{z}) \, \sim \, \mathcal{C}}[
\mathcal{S}(\tilde{y}, \tilde{z} | x)].
$$
After fine-tuning the student model on the corpus sampled from the teacher, to evaluate the model on a test instance $(x_{test}, y_{test})$ from the target task, we decode both a chain-of-thought $\tilde z_{test}$ and a predicted label $\tilde y_{test}$ from the student
%\jack{what decoding alg?}\footnote{\citet{} reports that ; we find the }, i.e., $\tilde z_{test}, \tilde y_{test} \sim \mathcal{S}(y, z | x_{test})$
and evaluate $\tilde y_{test}$ versus the true label $y_{test}$.
We consider two strategies for decoding. %This observation is closely related to ``self-consistency''~\citep{wang2022self}. There may be different valid chain-of-thoughts for a given question; 
% show that taking a majority vote over a number of chain-of-thought rationalization chains often outperforms, e.g., greedy decoding. Because there may be different valid chain-of-thoughts for a given question, we hypothesize that 
(1) Predict the most likely chain-of-thought and the label $\tilde z_{test}, \tilde y_{test} = \operatorname*{argmax}_{z, y} \mathcal{S}(z, y | x_{test}).$ This can be approximated by greedy decoding or beam search. (2) There may be different valid chain-of-thoughts for a given question and as a result, large language models distribute probability mass for a certain label across many diverse chain-of-thoughts~\citep{wang2022self}. Thus, it is beneficial to marginalize out the reasoning paths to find the most consistent answer: $\tilde y_{test} = \operatorname*{argmax}_{y} E_{z \sim \mathcal{S}(z| x_{test})} \mathcal{S}(y | z, x_{test}).$
This can be approximated by sampling multiple reasoning paths and take a majority vote among the predicted answers, dubbed ``self-consistency''~\citep{wang2022self}. We experiment with both approaches and conduct a discussion in \S \ref{sec:self_consistency}.

\section{Experiments}

\label{sec:sec_with_experiments}

We evaluate primarily on 3 target tasks: 1) CommonsenseQA (CSQA)~\cite{talmor2019commonsenseqa}, a 5-way multi-choice dataset; 2) OpenBookQA~\cite{OpenBookQA2018},  and 3) QuaRel~\cite{tafjord2019quarel}. While any model capable of few-shot chain-of-thought could be substituted, we use the \texttt{code-davinci-002} version of GPT-3\footnote{ \citet{wang2022rationale}  reports better CoT performance from this version compared to other GPT-3 models.} \cite{NEURIPS2020_1457c0d6} as our teacher model $\mathcal{T}$.
%\jack{is that right?} \yj{here, I'm curious about the analysis on size/version of teacher would be matter. footnote this?}\yj{do we need the date here? davinci-002 updates sometime, we can have it roughly, in appendix} 
We use OPT \cite{OPTLM} as our student model $\mathcal{S}$. %Both GPT-3 and OPT are auto-regressive left-to-right transformer language models, but the authors of OPT open-source their smaller model checkpoints
Our standard student model is OPT-1.3B (though we explore a range of student model sizes in \S\ref{sec:sec_with_model_and_dataset_scaling}). 

%We augment the human-authored chain-of-thoughts provided by~\citet{wei2022chain} and~\citet{wang2022rationale} with samples we curate to form our teacher prompts $\mathcal{P}$.
We sample from GPT-3 with a temperature of $T = 1.0$. For each training example, we sample $N=30$ rationales. 
OPT is fine-tuned with a batch size of 32 and a learning rate of $2\times10^{-5}$. %At inference time, we use greedy decoding on the student model unless specified otherwise (see \S\ref{sec:main_results} for discussions).
We use HuggingFace transformers \cite{transformers}, Pytorch \cite{NEURIPS2019_9015}, and Accelerate\footnote{\url{https://github.com/huggingface/accelerate}} for the implementation. Main experiments can be reproduced on one GPU with 48GB of memory.

\subsection{Results in Default SCoTD Setting}
\label{sec:main_results}
\label{sec:sec_with_supervised_results}
% In this section, we show that learning with GPT-3 generated rationales significantly improves the performance of small language models.

\begin{table}[t]
\caption{Performance before (a) and after (b) \our.}
\label{table:main}
\begin{center}

\begin{subtable}[c]{\linewidth}

\centering
\resizebox{\linewidth}{!}{
\begin{tabular}{ 
cc|ccc}
\toprule

Model & CoT & CSQA & QuaRel & OpenBookQA\\
\midrule

 \multirow{3}{*}{GPT3-175B} 
& No CoT &  \textbf{82.1} & \textbf{86.9} & 83.4 \\
&  Greedy & 77.6 & 83.3 & 71.8  \\
& Self-Consistency & 81.3 & 86.0 & \textbf{86.4}   \\

\midrule

\multirow{3}{*}{OPT-1.3B} & No CoT & 20.5 & 9.7 & 2.8  \\
& Greedy  & 17.9 &  39.6 & 12.6   \\
& Self-Consistency & 21.1 & 48.2 & 22.2   \\

\midrule
\textcolor{gray}{Random} & \textcolor{gray}{-} &  \textcolor{gray}{20.0}  & \textcolor{gray}{50.0}  & \textcolor{gray}{25.0} \\

%Human & - & - &  \\
\bottomrule
\end{tabular}
}

\subcaption{Performance of prompting the teacher (GPT3-175B) and student model (OPT-1.3B, before distillation). The student fails to outperform the random guess baseline.}
\end{subtable}

\vspace{3mm}
\begin{subtable}[c]{\linewidth}
\centering
\resizebox{\linewidth}{!}{
\begin{tabular}{ 
cc@{\hskip9pt}c@{\hskip9pt}c@{\hskip9pt} 
c@{\hskip9pt}c@{\hskip9pt}c@{\hskip9pt}
c@{\hskip9pt}c@{\hskip9pt}c@{\hskip9pt}}
\toprule

Labeled Data & CoT & CSQA & QuaRel & OpenBookQA\\
\midrule

% \multirow{3}{*}{Few-Shot} & \multirow{3}{*}{GPT-3} & \multirow{3}{*}{Prompt} 
% & No CoT & 82.1 & 86.9 & 83.4 \\
% &  & & Greedy CoT & 77.6 &   \\
% &  & & Majority Vote CoT & 81.3 & 86.0 & 86.4   \\

% \midrule

% \multirow{3}{*}{Few-Shot} & \multirow{3}{*}{OPT-1.3B} & \multirow{3}{*}{Prompt} & No CoT &   \\
% &  & & Greedy CoT &   \\
% &  & & Majority Vote CoT &   \\

% \midrule

\multirow{3}{*}{Few-Shot} &
 Label-Only & 62.7 & 65.6 & \textbf{59.8} \\
& Greedy-CoT & 64.6 & 64.7 & 48.8 \\
& SCoTD & \textbf{64.7} & \textbf{73.0} & 57.8 \\

\midrule

\multirow{3}{*}{Full}  & Label-Only & 63.0 & 59.0 & 60.2 \\
&  Greedy-CoT & \textbf{68.2} & 71.2 & 50.0 \\
&  SCoTD & 67.0 & \textbf{83.8} & \textbf{67.0} \\

%Human & - & - &  \\
\bottomrule
\end{tabular}
}
\subcaption{Performance of the the student model after distillation.}
\vspace{-2mm}
\end{subtable}

\end{center}
\end{table}

\begin{figure*} [t]
\centering
\includegraphics[width=\textwidth]{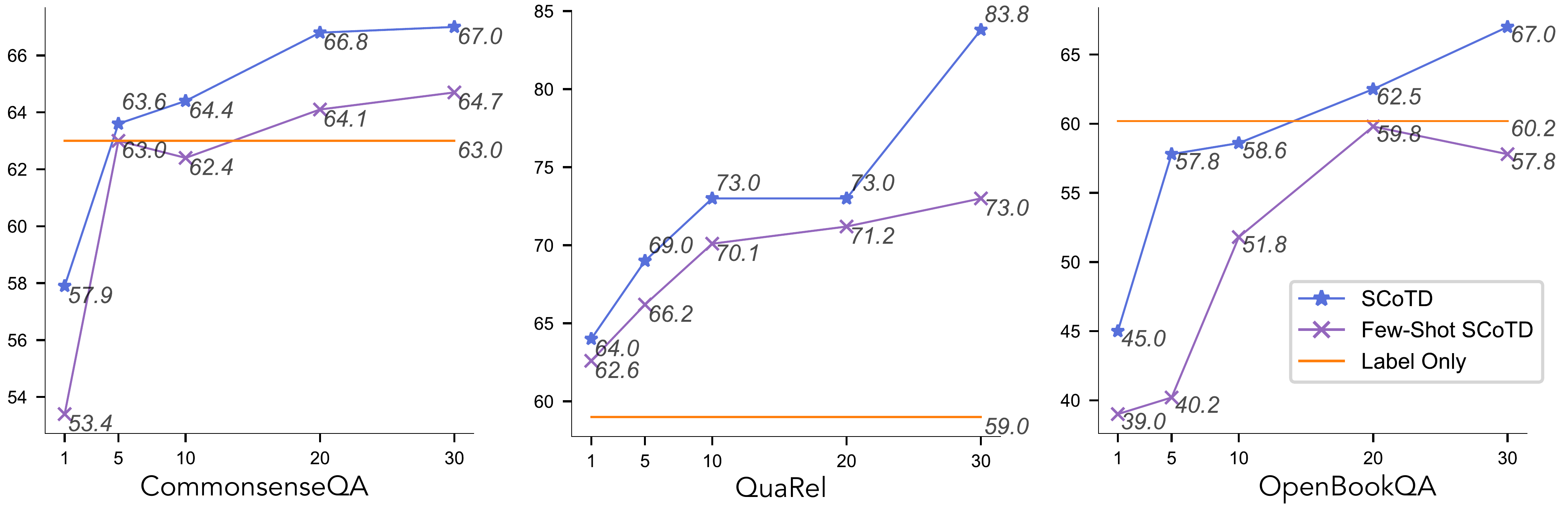}
\caption{For three commonsense QA tasks, accuracy (y-axis) improves significantly as the student is trained on more chain-of-thoughts sampled from the teacher (x-axis). Oversampling chain-of-thoughts is sometimes required to improve student performance beyond the supervised label-only baseline, e.g., as in OpenbookQA.
}
\label{fig:main_result}
\end{figure*}

% \paragraph{Few-shot SCoTD.}

We first consider both a few-shot learning setting and a supervised setting. For the few-shot setting, the only labeled examples available to our teacher/student models are contained in the prompt set $\mathcal{P}$ (but we use the unlabeled examples and teacher-generated chain-of-thoughts/labels for training).\footnote{In this setting, teacher samples can contain incorrect labels, thus preserving the few-shot nature of the task.} We also consider the supervised setting, where we assume access to labels in $\mathcal{D}_{\texttt{Train}}$. Supervised SCoTD involves simply discarding the samples within $\mathcal{C}$ that do not have the correct label prior to fine-tuning the student: for CommonsenseQA, OpenBookQA, and QuaRel, this results in discarding $40.4\%$, $45.0\%$, $34.2\%$ of chain-of-thoughts. For the few-shot setting, we decode with the self-consistency approach; for the supervised setting, we decode with greedy decoding (introduced in \S~\ref{sec:approach}; see an discussion in \S~\ref{sec:self_consistency}).

We compare SCoTD to 2 baselines: %1) $\mathcal{S}_{\text{No-CoT}}$, a student model that operates without chain of thoughts: we condition the student on the the labeled examples $(x_i, y_i) \in \mathcal{P}$ directly as in-context examples \jack{cite in context learning?};
1) \textbf{Label-Only}, the student is fine-tuned on just the label (in the few-shot setting, the label comes from the teacher and could be wrong; in the supervised setting, we use the gold label), instead of also with CoT;
2) \textbf{Greedy-CoT}, we decode a single-CoT per example (instead of $N=30$ samples) from $\mathcal{T}$ for each training example instead of sampling.
%and 3) \textbf{SCoTD}, the student model trained on the corpus $\mathcal{C}$ sampled from the teacher model with no additional filtering.
For additional reference, Table~\ref{fig:main_result} (a) reports the performance of the student (and teacher) in a variety of few-shot settings prior to applying any distillation: No CoT = few shot prompting with labeled instances from $\mathcal{P}$ but no $z_i$, Greedy and Self-Consistency are prompting with CoT but with different decoding strategies (\S~\ref{sec:approach}).

Table \ref{fig:main_result} (b) gives the performance of the student model after distillation in the supervised and few-shot settings. In all cases, distillation significantly improves the student model, and in all-but-one case, learning with CoT outperforms the label-only distillation baseline. While the student model initially fails to perform CoT through prompting (Table~\ref{fig:main_result} (a)) it learns to do so through distillation.

\paragraph{The number of samples.} 
In our default setting, to serve as our distillation corpus $\mathcal{C}$, we sample $N=30$ rationales from the teacher $\mathcal{T}$ for each (unlabelled) training instance.
% How does the model performance scale with different numbers of rationales per example?
Figure~\ref{fig:main_result} shows the performance of the student model when it is trained on corpora with fewer sampled CoT per instance:  % Compared to our default setting, 
% While our default experimental setting samples  rationales 
% Results suggest that
results suggest that learning with multiple sampled (albeit nosier) rationales/chain-of-thoughts per example is more beneficial than learning with one (most likely) rationale. Will more rationales bring more performance improvement? We sampled more rationales from GPT-3 to train the student model; however, this does not bring more performance gains. When $N=50$, the performance is similar to $N=30$: the model achieves 67.0 in accuracy on OpenBookQA (v.s. 67.0), 67.2 on CommonsenseQA (v.s. 67.0), 84.9 on QuaRel (v.s. 83.8). % Both the few-shot and supervised \our enjoy performance improvement as the number of samples increase; learning with chain-of-thoughts may not show its full potential until we scale up the chain-of-thought volume.

\subsubsection{Human Evaluations}

\label{sec:sec_with_human_experiments}

While SCoTD improves task accuracy significantly, we additionally conduct human evaluations to assess the generated chain-of-thoughts themselves (see Table~\ref{tab:example_CoTs} for samples). We sample instances from the CommonsenseQA, OpenBookQA, and QuaRel validation sets (300 instances per dataset), and conduct head-to-head human evaluations\footnote{We remove the final prediction from each chain-of-thought, and ask crowdworkers which is more coherent, fluent, and (importantly) likely to lead to a correct answer. We use Amazon Mechanical Turk and pay a minimum of \$15/hr, see Appendix~\ref{sec:sec_with_crowdworking_details} for more details, including a screenshot of the HIT.}
to assess:

\paragraph{Q1: Does SCoTD result in higher-quality chain-of-thoughts?} \emph{Test: OPT-1.3B versus OPT-1.3B + SCoTD.} Result: \textbf{Yes.} We assess this hypothesis on two subsets of instances: 1) a pure random sample (N=900); and 2) a set of instances for which both models eventually predicted the correct label (N=654). The second setting focuses more closely on the chain-of-thoughts themselves rather than the predictive accuracy of the model. SCoTD is superior in both settings: for the random sample setting, SCoTD won in 59\% of cases ($p$<.001), whereas in the correctness controlled setting, SCoTD won in 61\% of cases ($p$<.001). Results hold with $p<.05$ for each QA dataset individually.

\paragraph{Q2: Does a SCoTD student surpass the much larger teacher?} \emph{Test: OPT-1.3B + SCoTD versus text-davinci-002.} While the task accuracy of the teacher is still higher in most cases, \textbf{the student-generated CoT are comparable.}\footnote{See \S\ref{sec:limitations} for more discussion about the disparity between CoT-quality and task accuracy.} We again evaluate on: 1) a pure random sample (N=900); and 2) a correctness-controlled setting (N=659). The 100x smaller SCoTD's generations are competitive in both cases; we can't reject the null hypothesis of the crowd having equal preferences (OPT-1.3B + SCoTD wins in 47\% and 51\% of cases respectively, $p > .01$). Results hold for each dataset individually, as well.

\begin{table}[t]
\caption{Student performance with and without self-consistency. }
\label{table:consistency}
\begin{center}

\begin{subtable}[c]{\linewidth}
\centering
\resizebox{\linewidth}{!}{
\begin{tabular}{ 
cc@{\hskip9pt}c@{\hskip9pt}c@{\hskip9pt} 
c@{\hskip9pt}c@{\hskip9pt}c@{\hskip9pt}
c@{\hskip9pt}c@{\hskip9pt}c@{\hskip9pt}}
\toprule

Model & Self-Consistency & CSQA & QuaRel & OpenBookQA\\
\midrule

% \multirow{3}{*}{Few-Shot} & \multirow{3}{*}{GPT-3} & \multirow{3}{*}{Prompt} 
% & No CoT & 82.1 & 86.9 & 83.4 \\
% &  & & Greedy CoT & 77.6 &   \\
% &  & & Majority Vote CoT & 81.3 & 86.0 & 86.4   \\

% \midrule

% \multirow{3}{*}{Few-Shot} & \multirow{3}{*}{OPT-1.3B} & \multirow{3}{*}{Prompt} & No CoT &   \\
% &  & & Greedy CoT &   \\
% &  & & Majority Vote CoT &   \\

% \midrule

\multirow{2}{*}{Few-Shot \our} 
&
 No & 60.2  & 73.4 & 44.4 \\

& Yes & 64.7 (\textcolor{red}{+4.5}) & 73.0  (\textcolor{gray}{-0.4}) &  57.8 (\textcolor{red}{+13.4})\\
\midrule 

\multirow{2}{*}{\our} &
 No & 67.0  & 83.8 & 65.8 \\

& Yes & 66.8 (\textcolor{gray}{-0.2})  & 83.8 (\textcolor{gray}{-0.0}) & 63.6 (\textcolor{gray}{-2.2}) \\

%Human & - & - &  \\
\bottomrule
\end{tabular}
}
\subcaption{Self-consistency is most helpful under the few-shot setting, where we train with unfiltered and noisy CoTs. }
\end{subtable}

\vspace{3mm}
\begin{subtable}[c]{\linewidth}
\centering
\resizebox{\linewidth}{!}{
\begin{tabular}{ 
cc@{\hskip9pt}c@{\hskip9pt}c@{\hskip9pt} 
c@{\hskip9pt}c@{\hskip9pt}c@{\hskip9pt}
c@{\hskip9pt}c@{\hskip9pt}c@{\hskip9pt}}
\toprule

\multirow{2}{*}{Dataset} & \multirow{2}{*}{Self-Consistency} & \multicolumn{5}{c}{\#Rationales/Example}  \\
 & & 1 & 5 & 10 & 20 & 30 \\

\midrule

\multirow{2}{*}{CSQA} &  No & 53.0  & 58.3  & 59.1 & 60.0 & 60.2 \\

& Yes & 53.4 (\textcolor{red}{+0.4}) & 63.0 (\textcolor{red}{+4.7}) & 62.4 (\textcolor{red}{+3.3}) & 64.1 (\textcolor{red}{+4.1}) & 64.7 (\textcolor{red}{+4.5})\\

\midrule

\multirow{2}{*}{QuaRel} &  No & 62.2  & 68.7  & 69.8 & 70.9 & 73.4 \\

& Yes & 62.6 (\textcolor{red}{+0.4}) & 66.2 (\textcolor{gray}{-2.5}) & 70.1 (\textcolor{red}{+0.3}) & 71.2 (\textcolor{red}{+0.3}) & 73.0 (\textcolor{gray}{-0.4})\\
 
\midrule

\multirow{2}{*}{OpenBookQA} &  No & 39.0  & 40.2  & 40.6 & 43.2 & 44.4 \\

& Yes & 38.0 (\textcolor{gray}{-1.0}) & 37.6 (\textcolor{gray}{-2.6}) & 51.8 (\textcolor{red}{+11.2}) & 59.8 (\textcolor{red}{+16.6}) & 57.8 (\textcolor{red}{+13.4})\\

%Human & - & - &  \\
\bottomrule
\end{tabular}
}
\subcaption{Performance of Few-Shot \our with different numbers of sampled CoTs. Benefit of ``self-consistency'' is most prominent when training with multiple rationales per example on CSQA and OpenBookQA.}
\vspace{-2mm}
\end{subtable}

\end{center}
\end{table}

\begin{figure}[t]
\centering
\includegraphics[width=.8\textwidth]{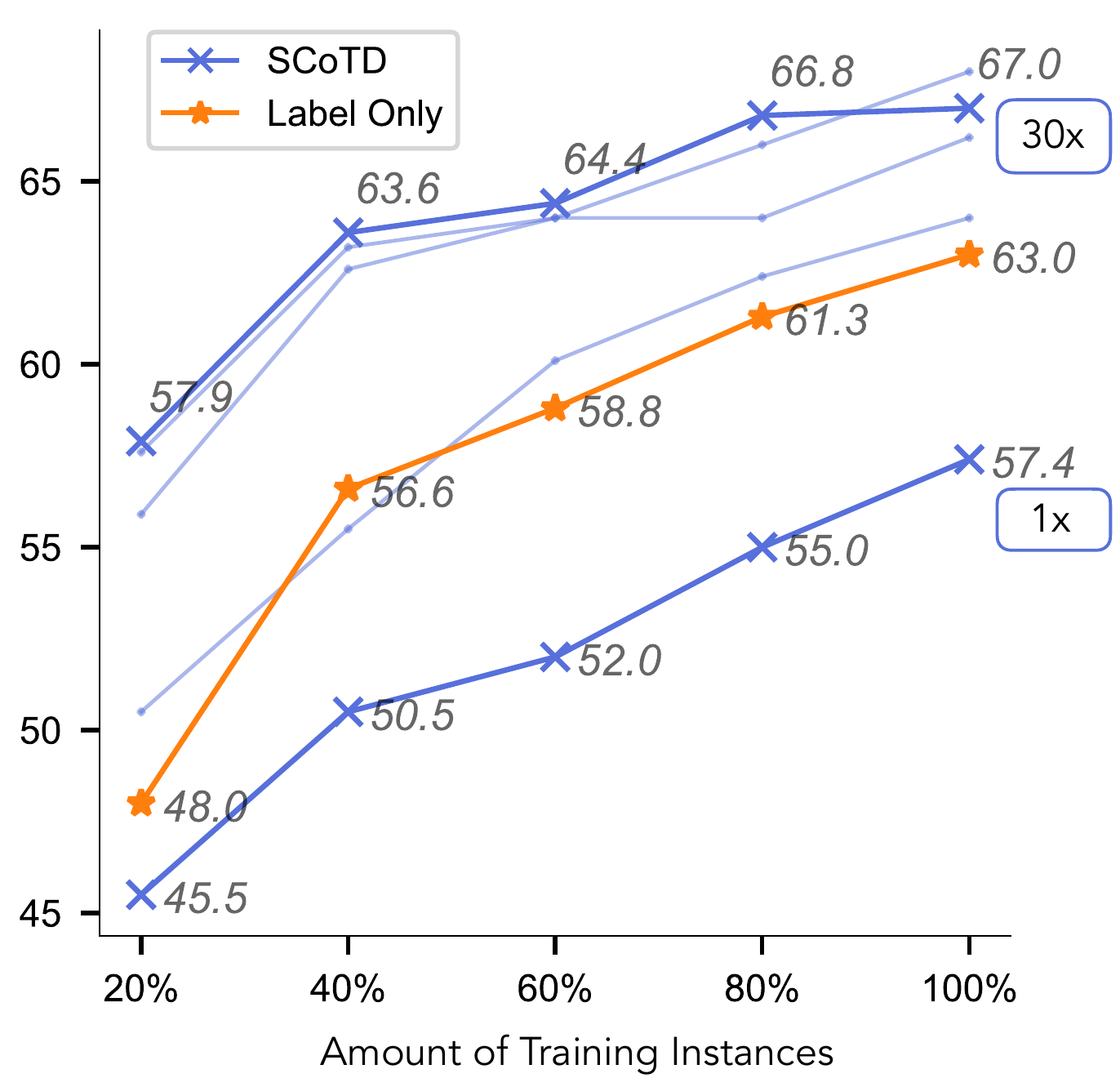}
\caption{Performance on CSQA with different amount of training instances, from using only $20\%$ of the $x$ from $\mathcal{D}_{\texttt{Train}}$ to using the full set (X-axis). Orange line is the Label Only baseline. Bottom blue line (marked with 1x) is \our but with only 1 sampled rationale per instance; above are \our with 5, 10, 20, 30 sampled rationales per instance, respectively.
}
\label{fig:data}
\end{figure}

\subsection{Self-Consistency for the Student}
\label{sec:self_consistency}

% This observation is closely related to ``self-consistency''~\citep{wang2022self}. There may be different valid chain-of-thoughts for a given question; 
% % show that taking a majority vote over a number of chain-of-thought rationalization chains often outperforms, e.g., greedy decoding. Because there may be different valid chain-of-thoughts for a given question, we hypothesize that 
% % We hypothesize that showing 
% and as a result, large language models may distribute probability mass for a certain label across many diverse chain-of-thoughts. Thus, during \textit{inference}, it is beneficial to marginalize out the sampled reasoning paths to find the most consistent answer. Specifically, instead of predicting the most likely chain-of-thought and the label $\tilde z_{test}, \tilde y_{test} = \operatorname*{argmax}_{z, y} \mathcal{T}(z, y | x_{test}, \mathcal{P})$ (approximated by greedy decoding or beam search), we marginalize over potential chain-of-thoughts: $\tilde y_{test} = \operatorname*{argmax}_{y} E_{z \sim \mathcal{T}(z| x_{test}, \mathcal{P})} \mathcal{T}(y | z, x_{test}, \mathcal{P}).$
% This can be approximated by sampling multiple reasoning paths and take a \textit{majority vote} among the predicted answers~\citep{wang2022self}.

% \paragraph{Self-consistency recovers performance when distilling with noisy CoTs.}
\citet{wang2022self} find that, for chain-of-thought prompted models, taking a majority vote over a large set of sample of predicted labels (resulting from a diverse range of CoTs) can improve performance. Our results regarding the effectiveness of sampling $N=30$ rationales from the teacher during SCoTD are similar-in-spirit: i.e., we also show performance gains from sampling multiple rationalization chains per instance.

% This explains why during distillation, it is beneficial to sample multiple chain-of-thoughts per instance from the teacher. 
% When we define our model to first predict the rationale and then a label, the true . While we could approximate the process 
% Indeed, \citet{wang2022self} term this phenomenon ``self-consistency'', where we approximate the marginalization with random samples, i.e., 
A natural question is, does the student model $\mathcal{S}$ exhibit the same phenomenon, i.e., can we sample multiple chain-of-thoughts from it and take a majority vote? We find that the student model can benefit from ``self-consistency,'' but not in all cases. In Table~\ref{table:consistency}, we report the performance with/without self-consistency (majority vote among 30 sampled reasoning paths with a temperature of $0.7$). When training with \textit{filtered} CoTs  (Table ~\ref{table:consistency} (a) bottom rows) or training with few CoTs per example (Table ~\ref{table:consistency} (b), when \#CoTs/Example is small), the student model does not benefit from self-consistency. Only when we train with multiple rationales per example without filtering (the few-shot setting), self-consistency is beneficial on CSQA and OpenBookQA. Overall, the results show that student models benefit from being shown a diverse/noisy set of rationales, and that self-consistency can be effectively applied after distillation.

\begin{figure} [h]
\centering
\includegraphics[width=.8\textwidth]{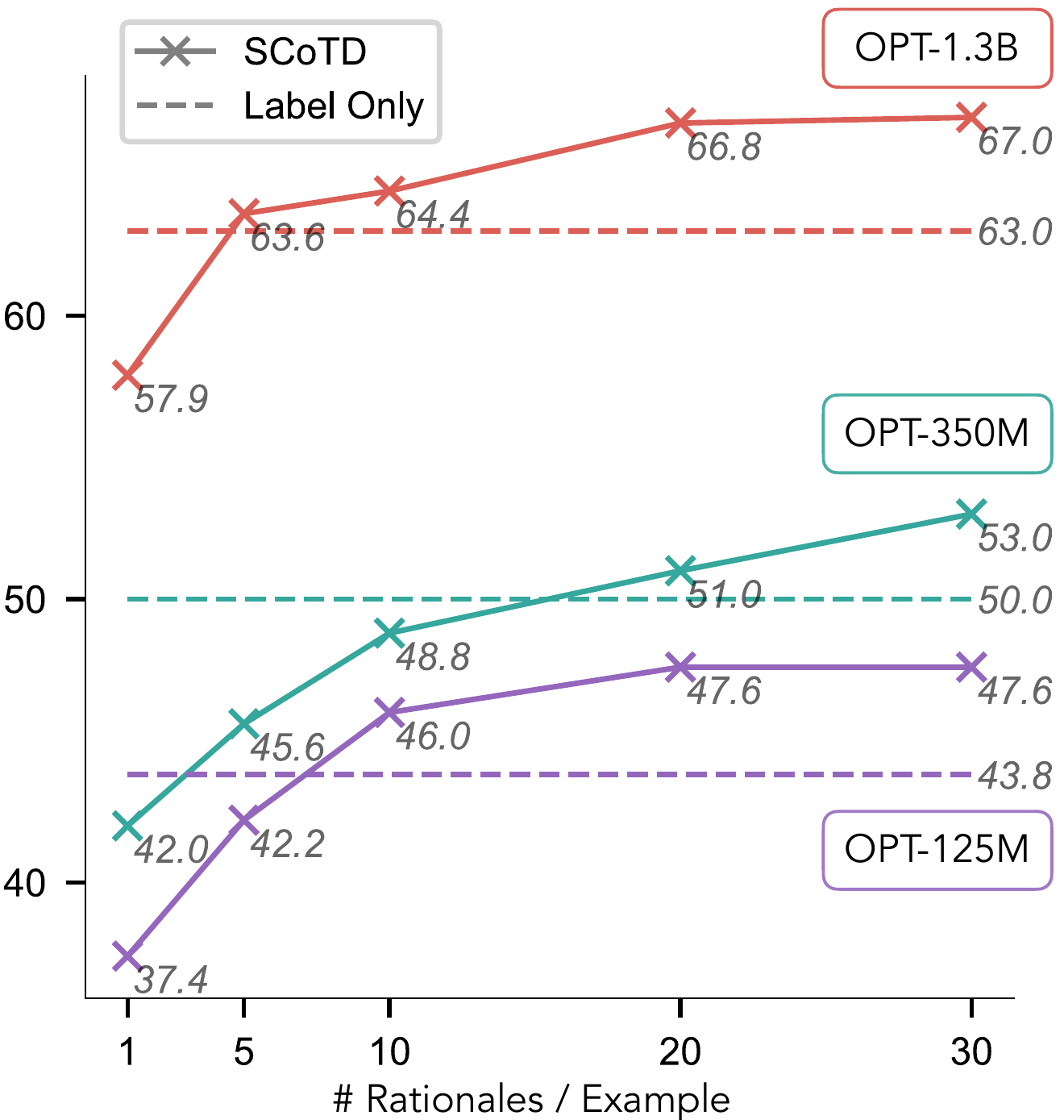}
\caption{Performance on CSQA with three different model sizes.
}
\label{fig:model_scale}
\end{figure}

\begin{figure*} [t]
\centering
\includegraphics[width=0.95\textwidth]{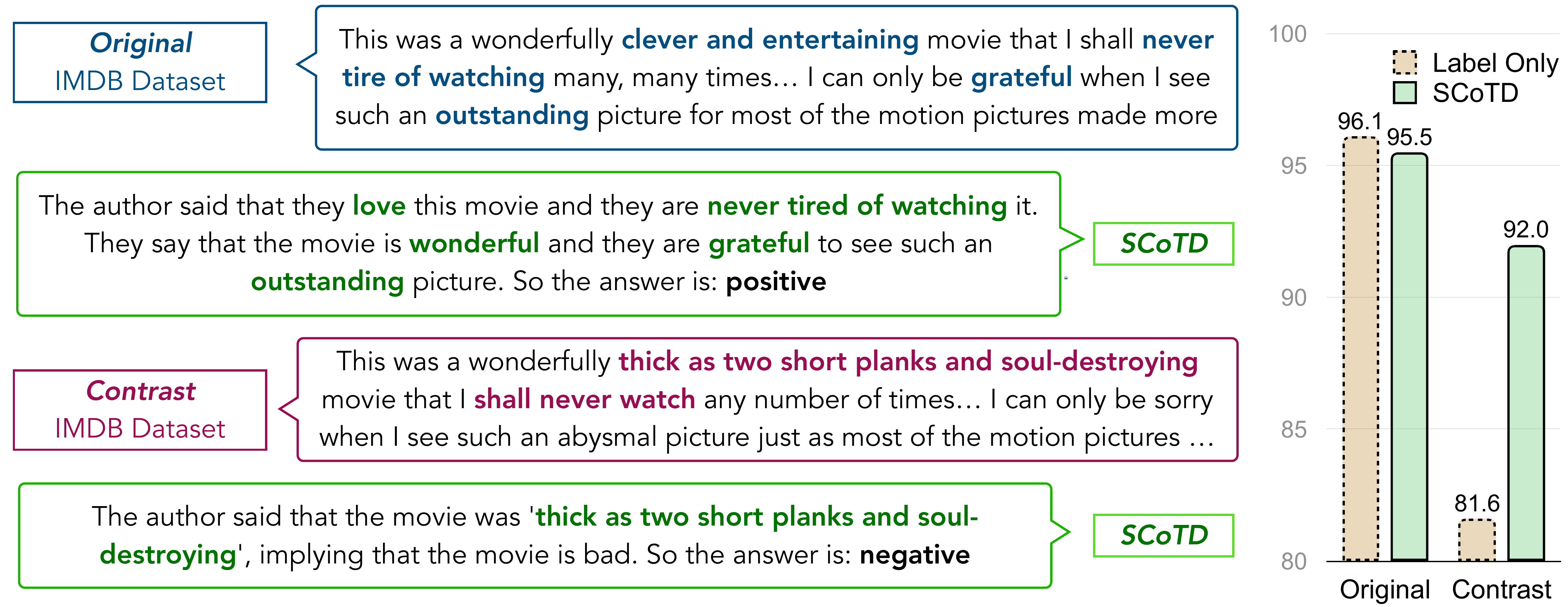}
\caption{Performance of SCoTD vs. label only supervision on the original and contrast IMDB dataset, along with sample predictions from SCoTD. %\yj{It would be nice if the failure sample of the base was shown, and also it would be nice if the main words of prediction and CoT were marked in bold.}, \harold{How about now?} %\yj{Among the training samples in imdb, I would like to create an example from various CoTs with various perspectives in training corpus.}
%\yj{looks great! we can save some space by moving original, contrast labels from left to right,}
}
\label{fig:imdb}
\end{figure*}

\subsection{SCoTD across Model and Dataset Sizes}

\label{sec:sec_with_model_and_dataset_scaling}

We also verify the effectiveness of \our across model and dataset sizes; in these experiments, we consider the supervised setting.

\paragraph{Data scaling.} Figure \ref{fig:data} shows the effect of varying the size of $\mathcal{D}_{\texttt{Train}}$ (for simplicity, we show only performance on CSQA as an example). Learning with CoTs is beneficial under all data scales. Interestingly, \our, trained with access to only $40\%$ of the labelled data, can surpass the direct supervised label-only model with $100\%$ of the labelled corpus; this result aligns with the argument in ~\citet{zaidan-etal-2007-using} -- providing more explanations from the teacher model could be more beneficial than providing more labels.

\paragraph{Student model size scaling.} Figure \ref{fig:model_scale} presents results when varying the size of the student model from 125M to 1.3B parameters for CSQA. For all model three model sizes, \our outperforms the standard supervised fine-tuning baseline (Label Only). Sampling multiple rationales per input instance is an effective strategy for all model sizes.

\subsection{SCoTD on Challenging Contrast Sets}

\label{sec:sec_with_contrast_sets}

Can learning with explanations help generalization, as hypothesized by~\citep{zaidan-etal-2007-using}? As a preliminary study, we show that \our enables better generalization to contrast sets. Contrast sets ~\citep{gardner2020evaluating} are proposed to evaluate a model’s robustness to perturbations around the decision boundary, by asking annotators to modify the original test instances in small but meaningful ways that (typically) change the gold label. %Contrast sets provide a local view of a model’s decision boundary. 

We experiment on the IMDB \cite{maas2011learning} sentiment analysis task in the supervised setting; we consider the corresponding contrast set of IMDB proposed by ~\citet{gardner2020evaluating}. We train two models on the training set of IMDB: \textbf{Label-Only} and \textbf{\our}. For efficiency, we sub-sample $100K$ examples from the training set of IMDB and truncate input sequences to 700 tokens. 
As shown in Figure \ref{fig:imdb}, while both models with/without \our achieve high performance on the original IMDB test set (96.1\% v.s. 95.5\%, with the Label-Only model performing slightly better), the model with \our achieves significantly higher performance on the contrast set: 92.0\% vs. 81.6\%. This result supports the hypothesis of ~\citep{zaidan-etal-2007-using}; that explanations can support more robust generalization.

\subsection{SCoTD on Unseen, Out-of-domain Tasks} 
% \yejin{'Out-of-Domain Tasks' might be a better phrasing than 'Fully-Held Out Tasks'...?} \harold{What about unseen tasks?}
\label{sec:sec_with_fully_held_out_task}

Large language models can perform few-shot, in-context learning with chain-of-thought prompting, i.e., generating reasonable chain-of-thoughts on \textit{unseen} tasks with a few demonstrations \cite{suzgun2022challenging}.
We conduct a preliminary experiment, inspired by \citet{min2021metaicl}'s MetaICL, to test whether student models trained with SCoTD acquire the same ability. We train a supervised \our model on ANLI, CommonsenseQA, and OpenBookQA, and evaluate it on SST-2 \cite{socher2013recursive}, a sentiment analysis task.

\begin{figure} [t]
\centering
\includegraphics[width=\textwidth]{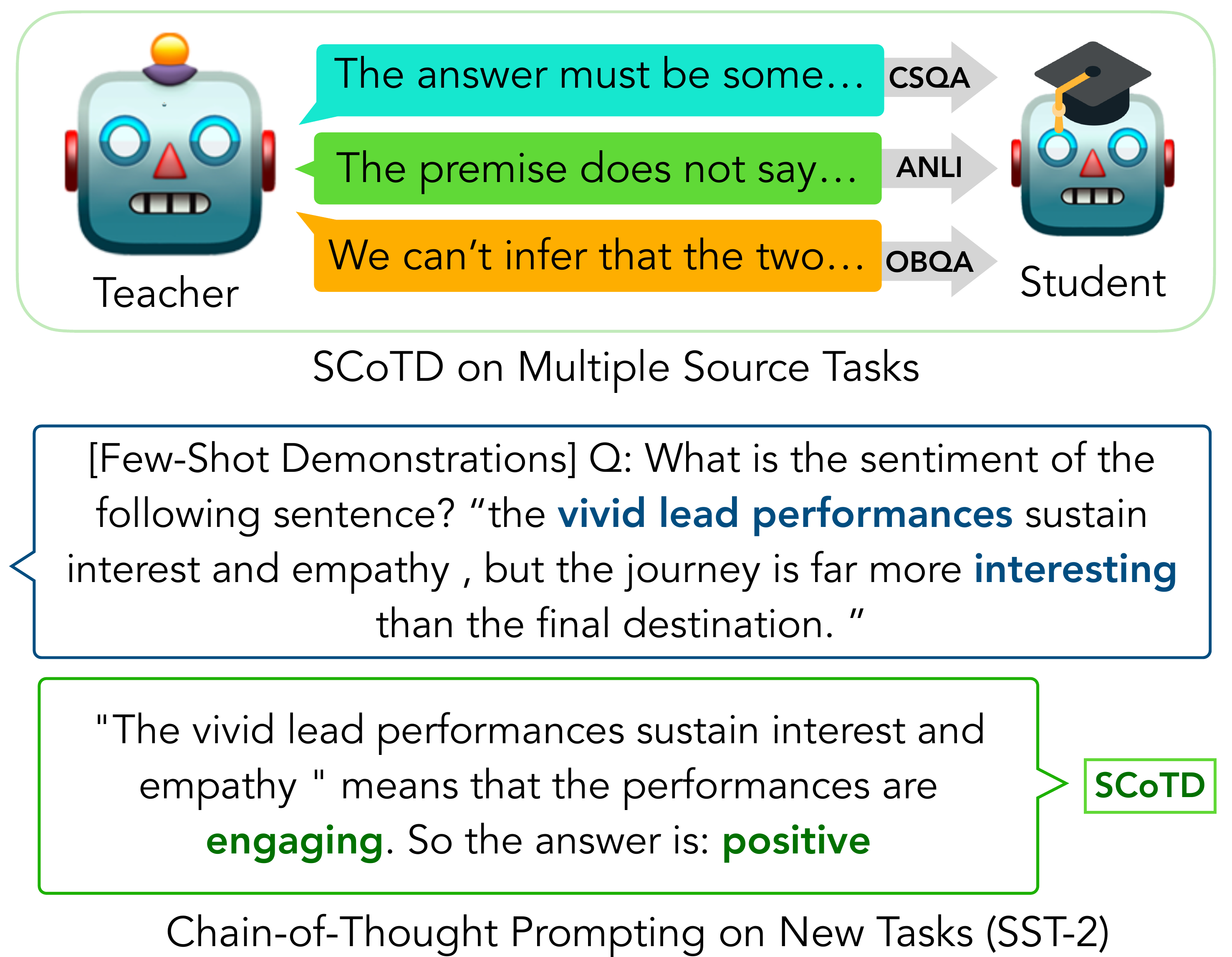}
\caption{Schematic of SCoTD models transferring from training tasks (CSQA, ANLI, OBQA) to unseen tasks (SST-2). %\yj{this is good! but it would be better if we unify font size and it's hard to read inside the green box, maybe we can size up the font fo readers to read easily}
}
\label{fig:transfer}
\end{figure}

The SCoTD model achieves a few-shot accuracy of $79.6\%$ on the validation set (an example prediction is shown in Figure \ref{fig:transfer}).\footnote{For reference, GPT-3 \texttt{text-curie-001} ($\sim$6.7B parameters) achieves $74.5\%$ with the same prompt.} %Intriguingly, we also experiment a
Compared to a baseline model that learns with no CoT% to do the standard in-context learning without the chain-of-thoughts
(i.e., a re-implementation of MetaICL trained on 3 source tasks); the baseline fails to recognize the input/output format of the new task and predicts answers out of the desired label set. It achieves (an effective)  $0\%$ accuracy on SST-2. %\jack{seems extreme... people might ask: why didn't you do the multiple-choice setting of scoring PPL of "positive" vs. "negative"?} \harold{That's a valid question... But it seems a bit too late to do this experiment.} \harold{I added a GPT-3 curie baseline; hopefully this puts the results in persepctive.} \jack{thank you!}
This suggests the potential of including CoTs during instruction/in-context tuning~\citep{wei2021finetuned,min2021metaicl}.

%We train a student model to learn how to perform in-context chain-of-thought reasoning on several source tasks: we maximize
%$
%E_{(x, \tilde{y}, \tilde{z}, \mathcal{P}) \, \sim \, \mathcal{C_{\mathcal{P}}}}[
%\mathcal{S}(\tilde{y}, \tilde{z} | x, \mathcal{P})],
%$ where $\mathcal{C}_{\mathcal{P}}$ consists of prompts and sampled chain-of-thoughts from multiple datasets. Then we transfer the student model to unseen tasks with a few human-authored chain-of-thought demonstrations. 

% As a preliminary study (as sampling more chain-of-thought rationales on dozens of datasets can be prohibitive),

\section{What Factors are Important for Distillation?}
\label{sec:sec_with_downsampling_experiments}

% \begin{table}[t]
% \caption{Ablation of .}
% \label{table:factor_analysis}
% \begin{center}
% \resizebox{0.8\linewidth}{!}{
% \begin{tabular}{ 
% cc@{\hskip9pt} c@{\hskip9pt}c@{\hskip9pt} 
% c@{\hskip9pt}c@{\hskip9pt}c@{\hskip9pt}
% c@{\hskip9pt}c@{\hskip9pt}c@{\hskip9pt}}
% \toprule

%  & QuaRel & OpenBook & CSQA\\

% \midrule

% Random & & 58.6 & 64.5 \\
% Diversity & & & 64.8 \\
% Probability & & & 65.1 \\
% \midrule
% 50 rationales &  \\

% \midrule

% %Human & - & - &  \\
% \bottomrule
% \end{tabular}
% }
% \end{center}
% \end{table}

\begin{figure} [t]
\centering
\includegraphics[width=.97\textwidth]{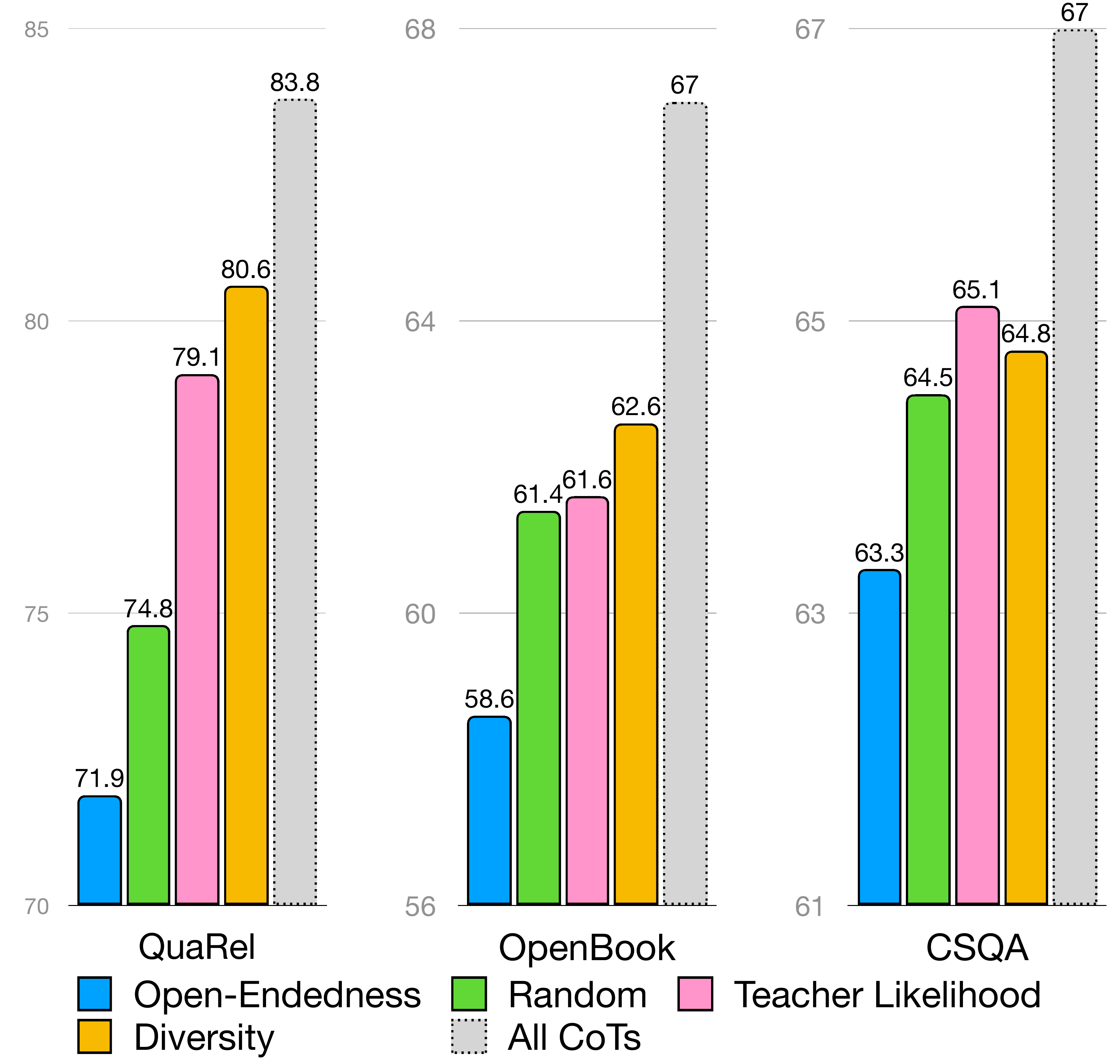}
\caption{Downsampling ablations: we subset our chain-of-thought distillation corpus $\mathcal{C}$ with a fixed budget according to different criteria. In general, keeping a diverse set of rationales performs well, though a random sample often performs well too.
}
\label{fig:downsample}
\end{figure}

%The experiments in \S\ref{sec:sec_with_experiments} show that SCoTD improves accuracy in a variety of settings compared to baselines, and that the student model can generate rationales of the same quality as the teacher.
An important factor underlying the performance gains highlighted in  \S\ref{sec:sec_with_experiments} was the number of chain-of-thoughts we sampled from the teacher model per-instance (more samples = better; Figure~\ref{fig:main_result}).
Here we ask: is data volume the key contributing factor to the performance improvement? Or, are specific aspects of chain-of-thought samples key for the performance improvements?

%Why does this process work, and what are the most important factors underlying its success? %A first observation (Figure~\ref{fig:main_result}) shows that as we sample more chain-of-thoughts from the teacher model, performance of the student model improves.

%\subsection{Ablation study}
% Beyond data volume, we explore several hypotheses regarding which factors are most important within chain-of-thought samples.

We design several filters to identify potentially important examples/CoTs among the correct rationales. We apply designed filters (to be introduced) to $\mathcal{C'}$, the corpus sampled from the teacher (with wrong CoTs dropped), that operationalize different hypotheses about what factors are important to distill. % We consider the few-shot setting for these experiments, i.e., some of the reasoning chains in $\mathcal{C}$ lead to the incorrect final answer. \jack{this is the few shot setting, right?} 
% We seek to use the filters to identify potentially important examples/rationales among the correct rationales. We first 
We control for dataset size when filtering, i.e., all filtered corpora have the same number of training CoTs. We downsample with a budget of 5 CoT per instance on average\footnote{In rare cases, we may end up with less as there are less than 5 correct CoTs for the instance.}. Then, we train the same student model on each of the filtered corpora, and compare on downstream tasks. If a student model trained on filtered corpus A tends to outperform the student model trained on filtered corpus B, then we argue that the property that produced corpus A is more important. The hypotheses we consider are:

\paragraph{Null hypothesis: data volume.} %Figure~\ref{fig:main_result} shows that, as we sample more chain-of-thoughts from the teacher model, performance of the student model tends to improve.
As a null hypothesis, we randomly sub-sample 5 CoT per instance; this filter operationalizes the assumption that an arbitrary set of samples is sufficient.

%\paragraph{Correctness.} Similar to our supervised setting, we sample 5 CoT that eventually lead to the correct prediction.

\paragraph{Diversity.} For each instance, we compute S-BERT \cite{reimers2019sentence} embeddings\footnote{We use \texttt{paraphrase-MiniLM-L6-v2}.} of each of the chain-of-thoughts, and cluster the resulting embeddings using hierarchical clustering into $k=5$ clusters. Then, we randomly sample a single instance from each cluster: the resulting sample covers all clusters, and thus represents a diverse+representative sample.

\paragraph{Teacher likelihood.} For each instance, we keep the 5 CoT samples with the highest per-token log-likelihood according to the teacher model.

\paragraph{Open-endedness.} Some instances in each dataset lead to a broader range of chain-of-thought samples than others. For example, on CommonsenseQA, the question ``What form of alcohol is made from grapes?" leads to a narrower range of rationalizations vs. ``Why might someone purposefully be going into trance?" 
We hypothesize that open-ended instances could benefit from relatively more sampled rationales.
We sort instances into quintiles based on the unique bi-grams in their corresponding 30 CoTs; for high-ranking instances (more unique CoT bi-grams, like the ``trance" example above), we keep more rationales and for low-ranking instances, we keep less rationales. We keep $1, 3, 5, 7, 9$ rationales for instances of different bins (thus controlling for the total number of CoT).

\paragraph{Results} Figure~\ref{fig:downsample} reports the accuracy of the student model when fine-tuned on the different subsampled corpora for the three tasks we consider. 
Overall, random subsampling is a strong baseline, but, we see some evidence that diversity among the rationales is important. None of the models trained on the sub-sampled data could approach the model trained on the full 30x/instance CoT set. This suggests that the sheer volume of the CoTs is a key driving force for the performance improvement.

% For \jack{which dataset?}, we continue this process beyond the initial set of 30 rationales to 50, and see diminishing returns, so: accuracy isn't likely to improve indefinitely just with sampling more.

% \paragraph{The Importance of Sampling Rationales}
% As we increase the number of sampled rationales from GPT-3, we not only get . We introduce a baseline where  

% \paragraph{Unsupervised Distillation} Is filtering (Sec \ref{}) necessary? We experiment with ``unsupervised distillation'', where we do not discard any rationales that are followed by a wrong answer. We use all the rationales sampled from GPT-3 without any filtering. This only results in a slight drop in performance.
 
% \paragraph{Scaling Up Number of Rationales} Does further sampling more rationales help? 

% \paragraph{GPT-3 v.s. Human Explanations}
% How well does human explanations work compared to 

%\section{Analysis}

%\paragraph{Model Size} Does rationalization benefit models of all sizes? We plot the 

% \paragraph{Dataset Size and the Number of Rationales} 

%\paragraph{Importance of Diverse Rationales}
%Why does scaling up the number of sampled rationales help and when should we stop sampling more rationales from the model? 

%The diversity of the rationales varies from example to example; for examples with diverse rationales, it is beneficial to sample more rationales. 

%One could use the n-gram diversity of the rationales as a proxy for the ``interestingness'' of the example. 

%\paragraph{Transfer to New Tasks}

\section{Related Work}

\paragraph{Chain-of-thought prompting.} As an extension of few-shot prompting \cite{NEURIPS2020_1457c0d6}, chain-of-thought has proven more generally applicable than algorithmic/structured reasoning for which intermediate step generation was initially studied, e.g., by \citet{roy2016solving,ling2017program,chiang2018semantically,https://doi.org/10.48550/arxiv.2112.00114}. Recent studies seek to improve and analyze CoTs from different perspectives: \citet{wang2022self} improves the original CoTs through marginalizing over diverse reasoning paths while \citet{wang2022rationale} marginalize over diverse prompts; \citet{zelikman2022star,huang2022large} improves CoT through a bootstrap manner of training on self-generated CoTs; \citet{li2022advance} introduce voting classifiers to filter sampled CoTs before final prediction; \citet{golovneva2022roscoe} introduce some automatic metrics for automatic assessment of chain-of-thoughts. This study instead focuses on enabling CoT for smaller models via distillation.
%  \paragraph{Self-consistency} \jack{harold must have a lot of citations for this like Meta-ICL etc.etc., could be merged into the above paragraph.}

\paragraph{Learning with explanations.}
% While the prevalent paradigm is to learn with annotated labels, there has been a long line of research on 
%\cite{deyoung2020eraser} a benchmark for evaluate rationale
\citet{hase2022can} discuss how explanations can serve as \emph{inputs} \cite{talmor2020leap}, \emph{targets} \cite{hendricks2016generating,fidler2017teaching,camburu2018snli,zhou2020towards,narang2020wt5,Kayser_2021_ICCV,wiegreffe2021reframing}, and \emph{priors} \cite{zhang2016rationale,srivastava2018zero} for machine learning models.
Chain-of-thought extends earlier efforts which treat explanations as intermediate structures, generated at inference time \cite{rajani2019explain}. Most related to our work is \citet{li2022explanations}, who do also learn with GPT-3 generated explanations; we show multiple samples improve significantly over their single-sample method, and also use chain-of-thought prompting at inference time vs. predicting explanations+labels via independent multitasking.

% \jack{need one sentence about how we differ, this one is 1 week outside of contemporaneous... worst case, it's contemporaneous with abstract deadline due to 13th lol}
%\cite{zhou2020towards}
%WT5 \cite{narang2020wt5}  % train the model to output the explanation after generating the prediction. 
%\cite{rajani2019explain} %Explain Yourself! Leveraging Language Models for Commonsense Reasoning,
%\citet{lampinen2022can} shows that explanations can improve in-context learning of large LMs; \citet{} explanations help supervising an image captioning model.
%explanations—plays a significant role in understanding abstract relational and causal structure~\cite{lampinen2022tell}
%E-ViL~\cite{}

\paragraph{Knowledge distillation.} Recent work, inspired by Knowledge Distillation~\cite{hinton2015distilling}, has considered symbolic knowledge distillation, \cite{west2021symbolic}, i.e., instead of distilling from soft representations like logits,
large language model serve as training data generators \cite{xiong2019pretrained,petroni2019language,schick-schutze-2021-generating,west2021symbolic,liu2022wanli,meng2022generating,bhagavatula2022i2d2}; this paper continues this line of work. %and uses large language models to generation CoTs/rationales (instead of input instances/output labels).

\paragraph{Contemporaneous work.} There are several contemporaneous papers: \citet{huang2022large}, \citet{magister2022teaching}, and \citet{ho2022large} all show that smaller models can benefit from large models' chains of thought.
We contributes beyond these by: 1) showing that sampling a large number of chain-of-thoughts is paramount; 2) exploring transfer performance to challenge sets/unseen tasks; and 3) analysis that address what factors are important in the teacher corpus.

\section{Conclusion}

We demonstrate the effectiveness of Symbolic Chain-of-thought Distillation (SCoTD): a method that enables smaller language models to effectively use chain-of-thought-style reasoning. We demonstrate the method's effectiveness across several downstream tasks, different student model sizes, different levels of supervision, and in difficult settings (challenge sets, unseen tasks). Our ablations shed light on what factors are particularly important to distill in these chain-of-thoughts. 

Our concrete recommendations are: 1) sampling multiple and diverse CoTs for each input instance, and 2) performing self-consistency when the teacher CoTs are noisy.
Several promising avenues for future work include:
\begin{enumerate}[leftmargin=*,topsep=1ex,itemsep=1ex,partopsep=1ex,parsep=1ex]
\item Exploring SCoTD for generation tasks in addition to classification tasks;
\item Scaling up the number of source tasks in \S~\ref{sec:sec_with_fully_held_out_task} to generalize to more tasks; %\jack{anyone have a good one for here? :)} \yj{applying chain of thought distillation beyond Text?} \yj{and I would like to see multi-lingual diverse chain of thought improve the final performance? related to limitations} \yj{training from multiple sources/experts, not only GPT3} ; and
\item Using the down-sampling setup introduced in \S\ref{sec:sec_with_downsampling_experiments} to explore additional hypotheses about what other factors may be of importance in CoTs.
\end{enumerate}

\section*{Limitations}

\label{sec:limitations}

Several limitations of our study include:
\begin{enumerate}[leftmargin=*,topsep=0pt,itemsep=-1ex,partopsep=1ex,parsep=1ex]
\item only English-language chain-of-thoughts/tasks considered;
\item reliance on GPT-3, which is a closed-source product with an unknown training set (which could itself include some explanations); and
\item focusing only on a single type of student model, OPT.
\end{enumerate}

More broadly, learning from and with explanations carries some specific risks related to automation bias. While a model might rationalize its predictions using a seemingly coherent string of natural language steps, even if it eventually gets the prediction correct, there's no guarantee that the eventually predicted output actually results from a process represented by the rationalization. A user might assign excessive confidence to that system based on the chain-of-thought. We observed many cases where the chain of thought seemed promising only to result in models ultimately making incorrect predictions in the final few tokens. Caution should be taken when displaying chain-of-thoughts to users.

\section*{Acknowledgment}
\label{sec:limitations}
We thank anonymous reviewers for their comments.
This work is supported in part by the DARPA MCS program, NCSOFT NLP Center and a Sloan research fellowship.

\bibliography{anthology,custom}
\bibliographystyle{acl_natbib}

\newpage

\appendix

\section{Crowdworking details}
\label{sec:sec_with_crowdworking_details}

\begin{figure*}
    \centering
    \includegraphics[width=.7\linewidth]{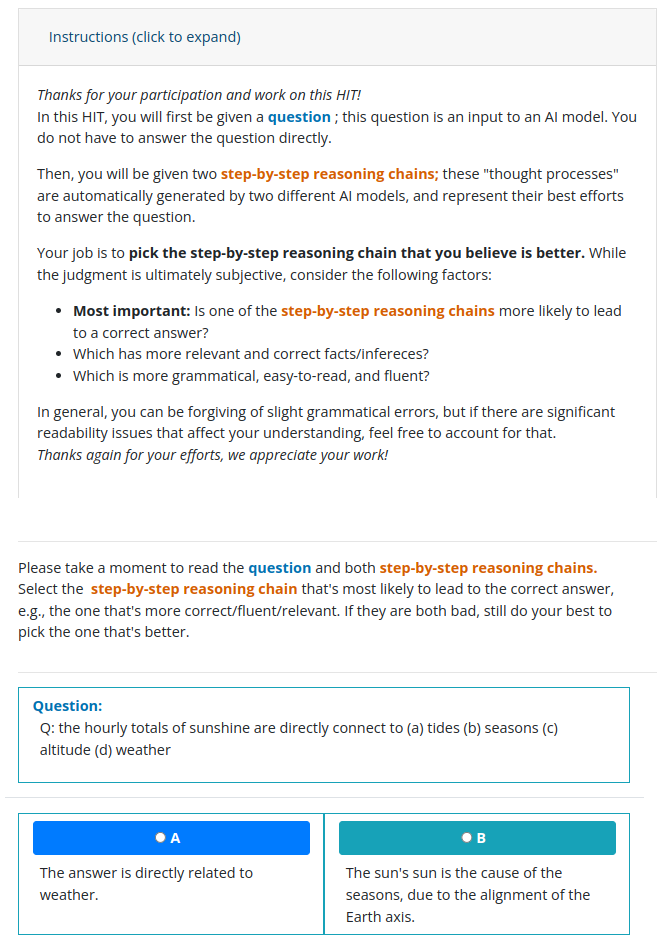}
    \caption{Crowdworking interface for pairwise judgements of chain-of-thought quality.}
    \label{fig:crowdworking}
\end{figure*}

A screenshot of the interface we use to collect the pairwise human judgments from \S\ref{sec:sec_with_human_experiments} is given in Figure~\ref{fig:crowdworking}. We conduct a post-hoc analysis using a javascript timer to ensure that annotators were paid at least \$15/hr: crowdworkers who didn't meet this hourly rate during annotation were awarded bonuses post-hoc to ensure they were paid that rate. We select crowdworkers with IP addresses in US,CA,NZ,AU,GB.

\paragraph{IRB Information} Crowdworking studies of standard NLP corpora (involving no personal disclosures) are not required by our IRB to be reviewed by them.
While the authors of this work are not lawyers and this is not legal advice,
this opinion is based on United States federal
regulation 45 CFR 46, under which this study qualifies as exempt. We do not release crowdworker IDs, so annotations cannot be back-traced to individual workers.

% \label{sec:appendix}

% This is an appendix.

\end{document}